\documentclass[10pt, a4paper]{article} 
\usepackage[english]{babel}
\usepackage[latin1]{inputenc}
\usepackage[T1]{fontenc} 
\usepackage{amsmath}
\usepackage{amssymb}
\usepackage{tikz}
\usepackage{calrsfs}
\usepackage{makeidx}
\usepackage{caption}
\usepackage{subcaption}
\usepackage{lipsum}
\usepackage{hyperref}
\usepackage{lettrine}
\usepackage{graphicx}
\usepackage{cleveref}
\usepackage{hyperref}
\usepackage{upgreek}
\usepackage{chemfig}
\usepackage{chemformula}
\usepackage{lipsum, parskip,libertine}

\usepackage{fancyhdr}
\usepackage{graphicx,type1cm,eso-pic,color}

\usepackage{ upgreek }
\usepackage{soul}

\usepackage{authblk}

\title{Modelling and characterization of  fine Particulate Matter   dynamics in Bujumbura using low cost sensors }

\author[1,2,4]{Egide Ndamuzi}
\author[2]{Rachel Akimana}
\author[3]{Paterne Gahungu}
\author[5]{Elie Bimenyimana}
\affil[1]{Doctorate School of the University of Burundi, Bujumbura, Burundi}
\affil[2]{University of Burundi, Bujumbura, Burundi}
\affil[3]{Imperial College London, Department of Mathematics, UK }
\affil[4]{Ecole Normale Sup\'{e}rieure de Bujumbura, Bujumbura, Burundi}
\affil[5]{Climate and Atmosphere Research Centre (CARE-C), The Cyprus Institute}

\date{} 

\begin{document}

\maketitle
	
	\section*{Abstract}
	Air pollution is a result of multiple sources including both natural and anthropogenic activities. The rapid urbanization of the cities such as Bujumbura economic capital of Burundi, is one of these factors. The very first characterization of the spatio-temporal variability of PM$_{2.5}$ in Bujumbura and the forecasting of PM$_{2.5}$ concentration have been conducted in this paper using data collected during a year, from august 2022 to august 2023, by low cost sensors installed in Bujumbura city.  For each commune, an hourly, daily and seasonal analysis were carried out and the results showed that
the mass concentrations of PM$_{2.5}$ in the three municipalities differ from one commune to another.  The average hourly and annual PM$_{2.5}$ concentrations   exceed the  World Health Organization standards. The range is between 28.3 and 35.0  $\mu$g/m$^3$. In order to make prediction of PM$_{2.5}$ concentration, an investigation of RNN with Long Short Term Memory (LSTM)  has been undertaken.


\section*{1. Introduction}
Air pollution remains one of the greatest environmental risks to global health and environment. Household combustion, motor vehicles, industrial facilities, waste burning and forest fires are common sources of air pollution. PM$_{2.5}$ can  enter the bloodstream, primarily resulting in cardiovascular and respiratory
diseases\cite{T1, T2, k1, 0k1, T3, CH1, CH2}. According to the W.H.O, there are 4.2 million premature deaths every year
as a result of exposure to ambient (outdoor) air pollution and 3.8 million for household (indoor)
exposure, primarily related to smoke from cooking and heating. Approximately 90 $\%$  of these deaths are estimated to occur in low and middle income countries, particularly sub-Saharan Africa. The estimates from W.H.O show that 80 $\%$ of people living in urban areas are exposed to air pollution
, threatening lives, productivity and economies.  
Efforts to  reduce air pollution across East African region have  been undermined by the lack of air quality oversight in key areas. A few reference monitors are deployed in some parts of sub-Saharan Africa. Air quality monitoring is currently guided by the use of low cost sensors in many regions\cite{k1}\cite{TT8}. Low cost particulate matter  sensors are becoming more widely available and are being increasingly deployed in ambient and indoor environments because of their low cost and ability to provide high spatial and temporal resolution PM information\cite{k08}.

\section*{2. Materials and Methods}	
\subsection*{2.1. Study area}
This study was carried out in Bujumbura (3$^{\circ}$ 21' 40.536'' S, 29$^{\circ}$ 20' 52.4976'' E; 774 m asl), largest and economic capital city of Burundi, located on the coast of Lake Tanganyika\textbf{Figure 1 (a)}. The city of Bujumbura is characterized by rather stable temperatures throughout the year ranging from 19 $^{\circ}$C to 29 $^{\circ}$C  and about 835mm of rainfall annually\cite{T003} \cite{rhugwasanye2022rainfall}. An annually rainfall peak have been found in the two months of April and May, where rainfall often reaches or exceeds 90 mm per month, and a minimum from June to August, where rainfall is rare and sporadic\cite{rhugwasanye2022rainfall}. Four seasons characterized the Bujumbura city: the long dry season (June-August), Short wet season (September to December), Short dry season (mid January-mid
February), and long wet season (February - May). Slight seasonal variations throughout the year characterize the average hourly wind speed in Bujumbura city: the windiest part of the year ( April-October)  and the calmest wind period of the year (October-April). As for wind direction, a predominant hourly average varies throughout the year in Bujumbura most often, one hand  from the south (January-February ; June-September) and other hand most from the east (February-June ; September-January)\cite{0TT4}.

Bujumbura city is one of the most densely populated cities in Burundi with density of 11, 668  inhabitant per km$^2$. The city's rapid urbanization has shifted disproportionately between the growth of population and urbanized area\cite{TT4}, which is associated with increase of air pollution levels due to anthropogenic activities. 

Three low-cost sensors have been installed in Bujumbura at different locations within the city. These three sites were selected to represent the key environments and main sources of air pollution in the city of Bujumbura, such as residential areas, industrial zones and business centres. The LCS were installed  in urban commune  \textbf{Mukaza} (3$^{\circ}$ 19' 12'' S, 29$^{\circ}$ 22' 19.2'' E), in \textbf{Ntahangwa} (3$^{\circ}$ 20' 57.768'' S, 29$^{\circ}$ 21' 0'' E) and in urban commune \textbf{Muha } (3$^{\circ}$ 26' 24'' S, 29$^{\circ}$ 22' 19.2'' E) \textbf{Figure 1 (b)}.

\begin{figure}[!ht]
     \centering
     \begin{subfigure}[b]{0.4\textwidth}
         \centering
         \includegraphics[width=7cm,height=70mm]{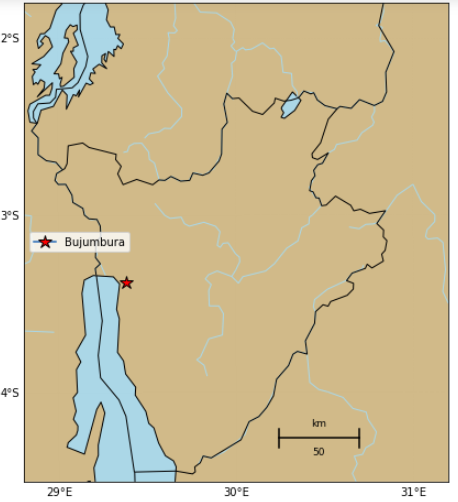}
        \caption{\textit{ Location of Bujumbura city in Burundi }}
        \label{F1a}
     \end{subfigure}
     \hfill
     \begin{subfigure}[b]{0.4\textwidth}
         \centering
         \includegraphics[width=7cm,height=70mm]{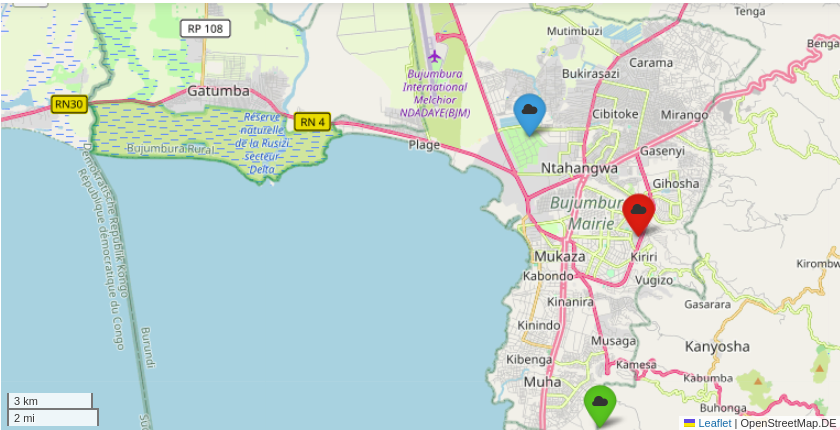}
        \caption{\textit{PM measurement sites in Bujumbura } }
         \label{F1b}
     \end{subfigure}
        \caption{The map of Burundi showing the location of the city of Bujumbura in Burundi (a) and PM measurement sites in Bujumbura (b)}
        \label{F1}
\end{figure}

\subsection*{2.2. Instrumentation}
Particulate matter measurements were conducted for a complete year (August 2022 to August 2023) in order to capture seasonal variability of PM concentrations in Bujumbura. PM concentrations were monitored by PurpleAir air quality sensors (PA-II-FLEX)\cite{TT08}. Plantower PMS 5003 optical sensors and a Bosch BME280 sensor are used  to estimate PM2.5 mass concentrations and  temperature, relative humidity, and pressure respectively\cite{T2}\cite{0TT08}. The functionality of these devices allows transmission of data  via
wireless connectivity in real time and recorded to an on-board micro SD card.  The devices measure sensor readings in six size bins ranging from 300 nm to 10 $\mu$ m  at approximately a 60-s interval. A proprietary algorithm, sometimes CF=ATM algorithm, converts raw sensor measurements to PM mass using assumptions about particle shape and density\cite{T2}.

\subsection*{2.3. Data correction}
Despite the availability of low cost particle sensors, their limited accuracy means that the data cannot be used reliably without correction. The common method used for data correction (calibration) is to co-locate the LCS with reference monitors\cite{Li2020}. Due to absence of reference instruments, we opted to use a transparent and reproducible alternative method ALT CF3 of calculating PM$_{2.5}$ from the number of particles in three size categories. Note that this approach was applied successfully in a couple of studies\cite{AC1, AC2}. Basically, the PM  concentrations data collected by PurpleAir low cost sensors is corrected using he particles concentrations in three size bins as expressed in equation \ref{size}

\begin{equation}\label{size}
    \text{PM}_{2.5}= 3(\alpha \text{X }+ \beta \text{Y} + \gamma \text{Z}),
\end{equation}
where the number 3 is the calibration factor(CF) \cite{AC1}. X, Y and Z are particle numbers per deciliter in three size categories, i.e X=0.3 $\mu$m - 0.5 $\mu$m, Y=0.5 $\mu$m - 1.0 $\mu$m and Z=1.0 $\mu$m - 2.5 $\mu$m, given directly by the PurpleAir LCSs. The estimate mass concentration in each size category, $\alpha, \beta$ and $\gamma$ respectively is given as a product of water density ($1g/cm^3$), the particle volume $\left(V=\frac{4}{3} \pi r^3 \right)$ and the number of particle (N); $r=\frac{d}{2}$ with $d$ the geometric mean of each size boundaries. It can also be approximated by the midpoint. Hence, $\alpha=0.00030418, \beta=0.0018512$ and $\gamma=0.02069706$.

\subsection*{2.4. Mathematical modelling}
Linear models, sometimes are not sufficient to capture the real-world phenomena, thus nonlinear
models are necessary\cite{k10}. The artificial neural networks (ANNs) are among of nonlinear
models used for nonlinear regression and classification tasks. Neural networks architectures have been widely used to perform forecasting tasks . In some cases, Neural Networks are viable competitors for various traditional time series models\cite{k1, k7, k9}. A recurrent neural network is  adapted to work for time series data or data that involves sequences. RNNs have the concept of memory that helps them store the states or information of previous inputs to generate the next output of the sequence i.e in RNN, the information cycles through the loop in the hidden layer such that the states information derived from earlier inputs remains in the network \cite{k11}.
The input layer at time step $t$, $x_t \in \mathbb{R}$ such that we can extend it to d-dimensional feature vector. Then, $x_t$ processes the initial input and passes it to the middle layer $h_t$ which consists of multiple hidden layers, each with its activation functions, weights, and biases. These parameters are standardized across the hidden layers so that instead of creating multiple hidden layers, it will create one and loop it over. The  rolled RNN represents the whole neural network, or rather the entire predict phase. The unrolled or unfolded RNN represents the individual layers, or time steps, of the neural network. In simplest terms, the equations \ref{e1} and \ref{e2}  define how an RNN evolves over time. The output $y$ at time $t$ is computed as

\begin{equation}\label{e1}
y_t=f(h_t, w_y)=f(w_y . h_t +b_y).
\end{equation}
where $.$ is the dot product, $w_y \in \mathbb{R}^m$ are weights associated with hidden to output units with $m$ number of hidden layers, and $b_y \in \mathbb{R}$  is the bias associated with the feedforward layer. The RNN can remember its past by allowing past computations $h_{t-1}$ to influence the present computations $h_{t}$.
\begin{equation}\label{e2}
h_t=g(x_{t}, h_{t-1}, w_x, w_h, b_h)=g(w_x x_{t}+w_h h_{t-1}+b_h),	
\end{equation}
where $w_x \in \mathbb{R}^m$ are weights associated with inputs in recurrent layer and $b_h \in \mathbb{R}^m$ is the bias associated with the recurrent layer. $f$ and $g$ are activation functions. In the recurrent neural network, any activation function we like in $t$ time can be used.

Long Short Term Memory(LSTM) are a special kind of RNN, capable to overcome the vanishing/exploding gradient problem, so RNNs can safely be applied to extremely long sequences. The accuracy of the model is based on the metrics such as root mean square error(RMSE) 

\begin{equation}
RMSE=\sqrt{\frac{1}{n} \sum_{i=1}^n (y_i - \hat{y}_i)^2}
\end{equation}
and absolute mean error(AME)

\begin{equation}
MAE=\frac{1}{n} \sum_{i=1}^n |y_i - \hat{y}_i|
\end{equation}
where $n$ denotes the total number of observations, $y_i$
denotes the observed values, and $\hat{y}_i$ denotes the predicted values.

\section*{3. Results and Discussions}
The first-ever characterization of the spatio-temporal variability of PM$_{2.5}$ in Bujumbura using low-cost sensors between August
2022 to August 2023 revealed that PM$_{2.5}$ mass concentrations in the municipalities of Bujumbura differ from one commune to another.  A very high PM2.5 annually mean concentration was observed in Muha  (35.0 $\mu$g/m$^3$), and the lowest in Ntahangwa  (28.3 $\mu$g/m$^3$).  For Mukaza, 32.8 $\mu$g/m$^3$ PM$_{2.5}$ annually mean concentration was observed.  
\subsection*{3.1. Comparison between CF1 and ALT CF3 algorithms}
Plantower algorithm (CF1) was compared with the ALT CF3 approach for the three sites (Ntahangwa, Muha and Mukaza) and the corresponding results are shown in \textbf{Figure 2}. Overall good agreement between the two algorithms was observed in terms of temporal trends as demonstrated by strong correlations between the two methods. However, higher concentrations were observed for CF1 compared to ALT CF3 (almost two times higher). This is in accordance with what is reported in previous studies where an overestimate of PM$_{2.5}$ concentrations of approximately 40 $\%$ were found for the PurpleAir LCSs compared to the reference instruments\cite{AC1} \cite{AC4} \cite{Li2020}. In fact, estimates provided by Plantower\cite{AC0}, the manufacturer of the sensors used in PurpleAir monitors using the (CF1 or CF-ATM) method was inferior in several respects\cite{0k1} \cite{k08}  (lower precision, higher detection limit, less improved size distribution), PM$_{2.5}$ concentrations overestimation  compared to this alternative. Hence, the precision of LCSs is still not comparable to reference-grade measurements\cite{AC3}.

\begin{figure}[!ht]
     \centering
     \begin{subfigure}[b]{0.3\textwidth}
         \centering
         \includegraphics[width=\textwidth]{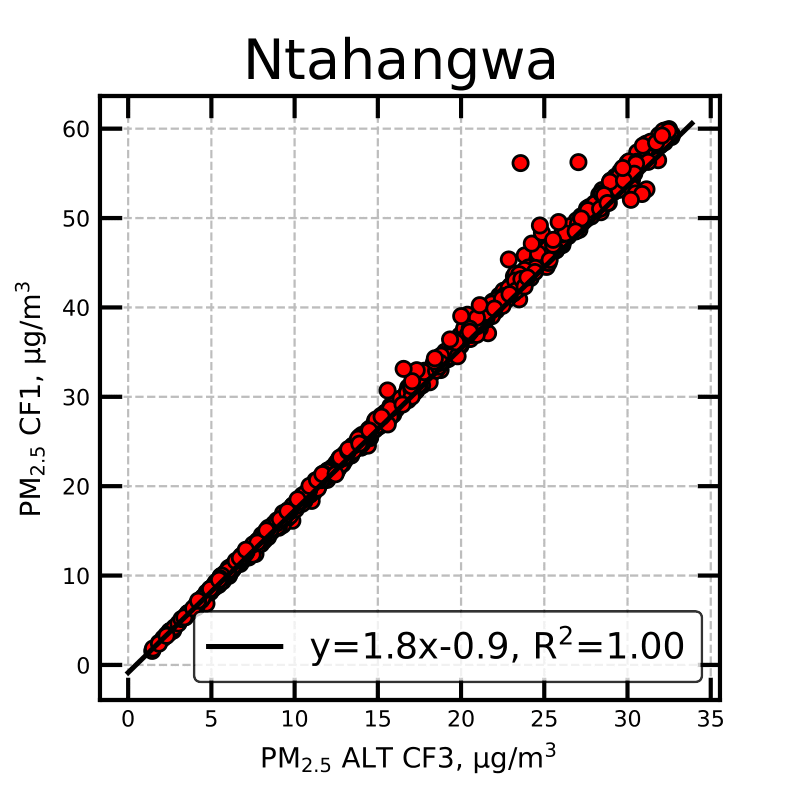}
         \label{fig:y equals x}
     \end{subfigure}
     \hfill
     \begin{subfigure}[b]{0.3\textwidth}
         \centering
         \includegraphics[width=\textwidth]{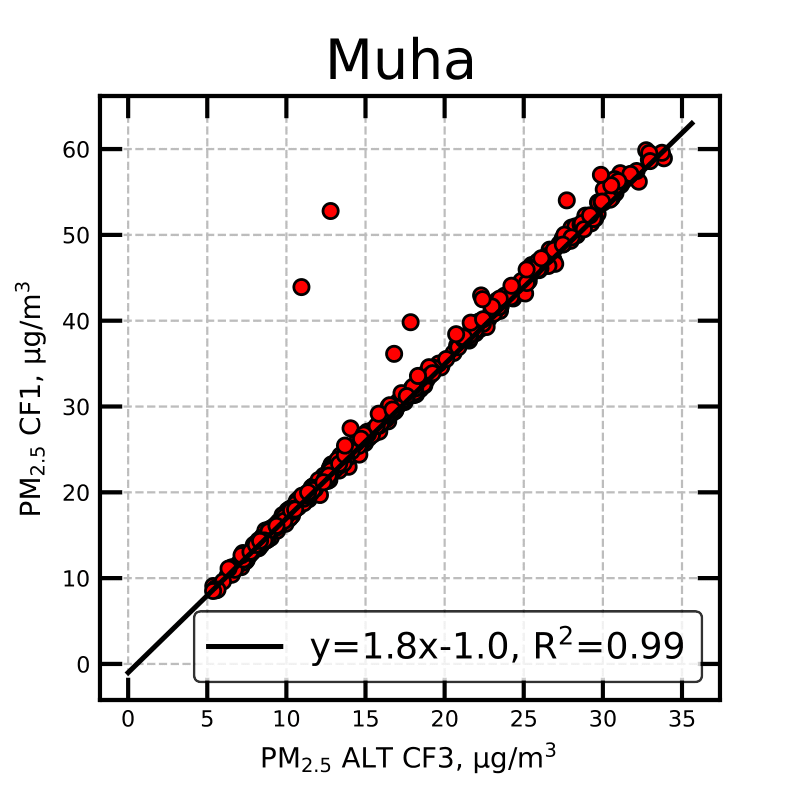}
         \label{fig:three sin x}
     \end{subfigure}
     \hfill
     \begin{subfigure}[b]{0.3\textwidth}
         \centering
         \includegraphics[width=\textwidth]{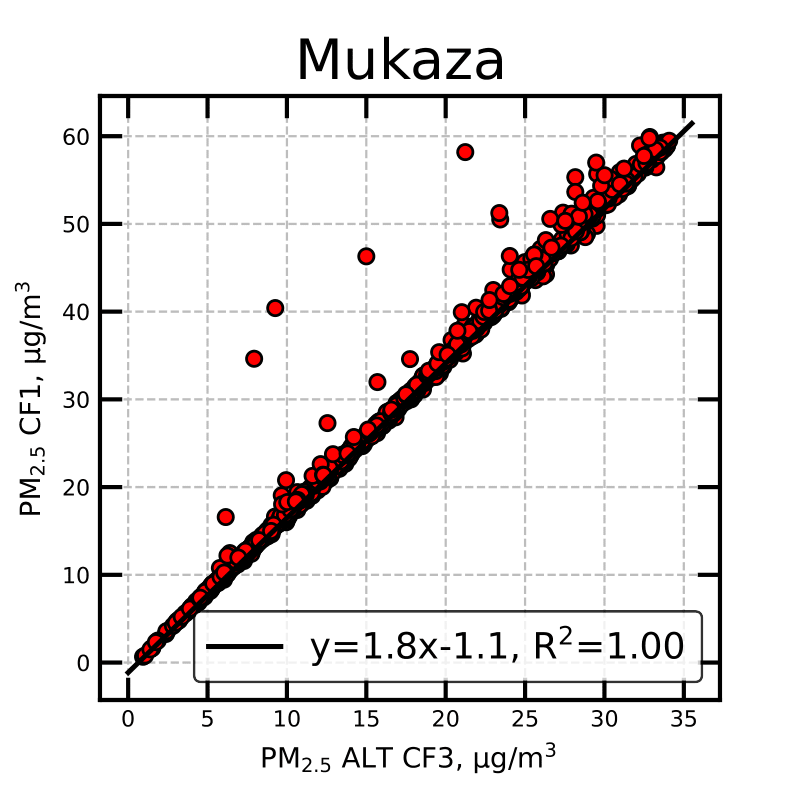}
         \label{fig:five over x}
     \end{subfigure}
        \caption{ALT CF3 algorithm vs CF1 Plantower algorithm}
        \label{fig:three graphs}
\end{figure}

\subsection*{3.2. Diurnal variability of PM$_{2.5}$}
Diurnal cycles of PM$_{2.5}$ concentrations across the three sites are shown in \textbf{Figure 3} where the error bars represent the standard deviation. Mean hourly PM$_{2.5}$ mass concentrations of  32.9 $\mu$g/m$^3$, 24.2 $\mu$g/m$^3$ and 31.4 $\mu$g/m$^3$  have been observed in Mukaza, Ntahangwa and Muha respectively.

\begin{center}
	\begin{figure}[!ht]
		\centering
		\includegraphics[scale=0.6]{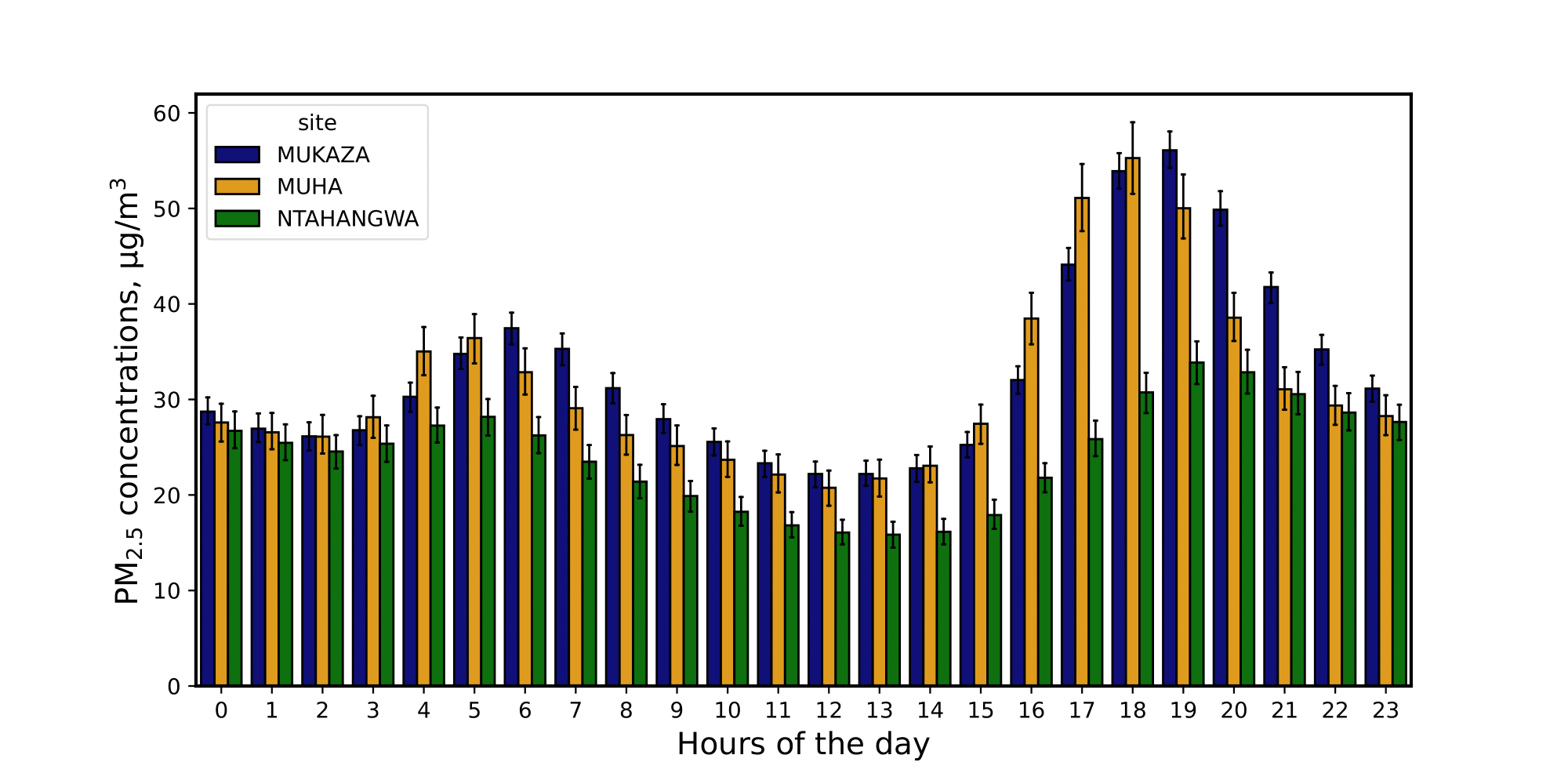}
		\caption{Hourly mean value of PM$_{2.5}$ mass concentrations in Bujumbura}
	\end{figure}
\end{center}
As shown in \textbf{Figure 3}, two peaks of PM$_{2.5}$ were observed at the three sites in the morning and evening. In general, the city of Bujumbura has traffic in the form of a transport network oriented towards the city center which is particularly noticeable during rush hours. For the Muha and Ntahangwa, a PM$_{2.5}$ morning peak  can generally be due to the road traffic. The population of these two municipalities often go at time to the city center, i.e towards Mukaza, a center considered as of business, commerce and administration. For the Mukaza, the PM$_{2.5}$ morning peak  can also be associated with the road traffic, i.e the arrival of those who come from these two other municipalities.  According to the study carried out by JICA (Japanese International Cooperation Agency)\cite{rep}, 80 $\%$ of vehicles leave the outskirts every morning heading towards the Bujumbura city center.

 After the morning peaks, the concentration of PM$_{2.5}$ decreases. The other peak is observed towards the evening which can also be associated with traffic, the return of those who have been at work and domestic activities such as cooking. Weather conditions such as wind direction could also explain the observed PM$_{2.5}$ diurnal variation given that Bujumbura is a coastal city and is likely to be affected by sea breeze and land breeze phenomena. 

\subsection*{3.3. Seasonal variability of PM$_{2.5}$}
The seasonal variability of PM$_{2.5}$ mass concentrations is shown in \textbf{Figure 4}. It is shown that the PM$_{2.5}$ mass concentrations increased significantly during the dry season and decreased during the rainy season. Several factors may explain this phenomenon. During the dry season, there is an increase in re-suspended dust as a result of dry surface and higher wind speed, as well as other human activities such as bush fires that are likely to occur during dry period. The slight increase in PM$_{2.5}$ mass concentrations observed around February-March is probably due to burning of agricultural residues since at that time people are preparing their fields for the main growing season.

\begin{center}
	\begin{figure}[!ht]
		\includegraphics[scale=0.8]{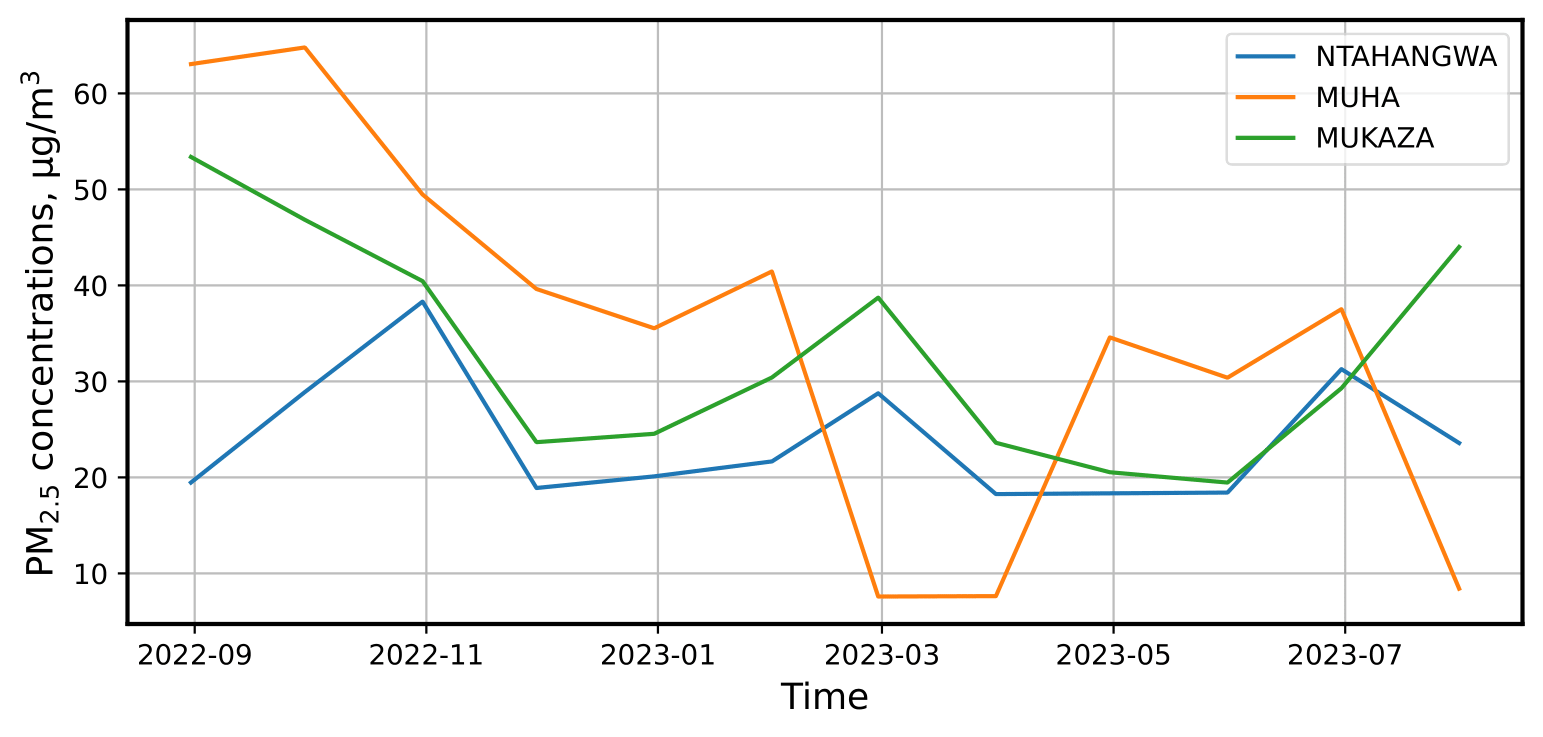}
		\caption{Seasonal variability of PM$_{2.5}$ mass concentrations in Bujumbura}
	\end{figure}
\end{center}

\subsection*{3.4. Daily mean PM$_{2.5}$  concentrations}

\begin{center}
	\begin{figure}[!ht]
		\includegraphics[scale=0.6]{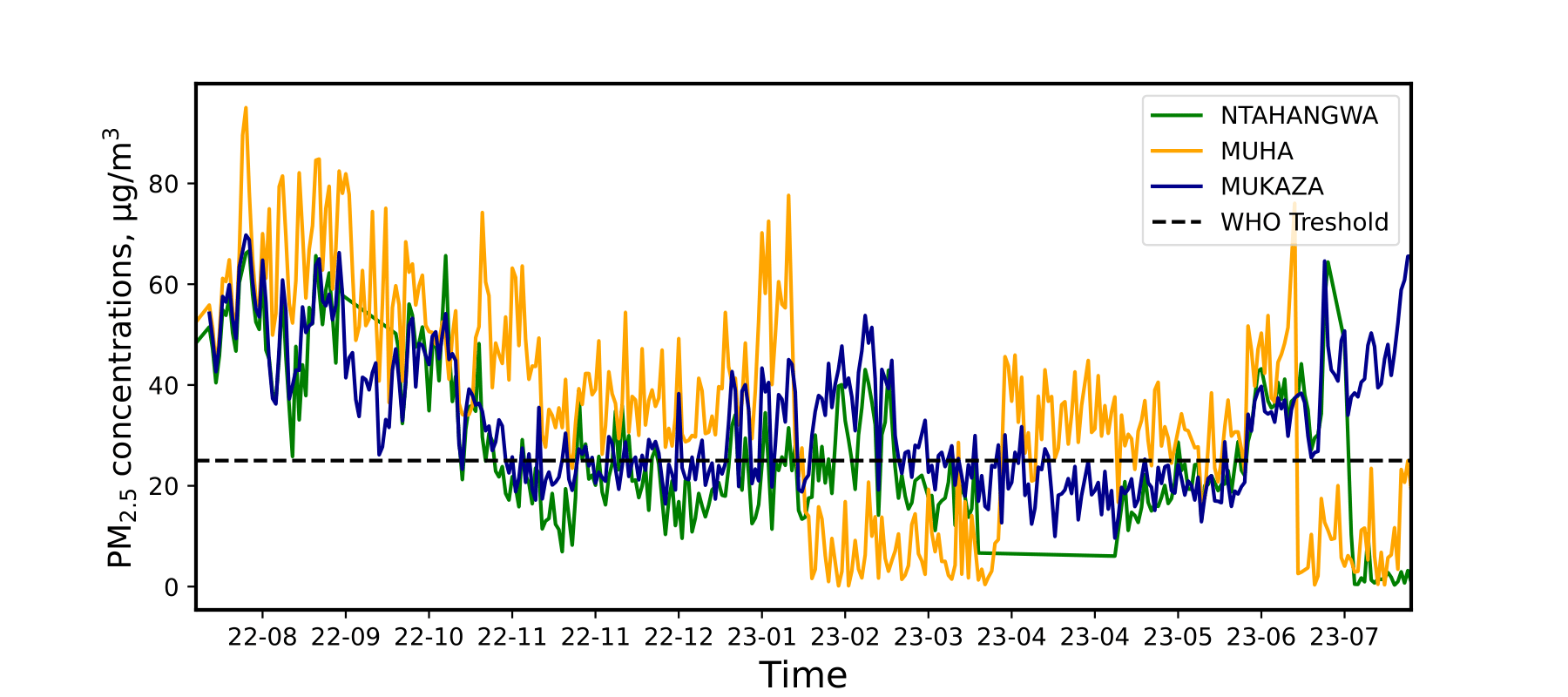}
		\caption{Daily variability of PM$_{2.5}$ mass concentrations in Bujumbura}
	\end{figure}
\end{center}
As shown in \textbf{Figure 5}, the mean daily mass concentrations of PM$_{2.5}$ indicates that  WHO guidelines of 25 $\mu$g/m$^3$ as a 24-hour mean  was exceeded in all municipalities of Bujumbura during the dry seasonal. Considering the PM$_{2.5}$ annually mean mass concentrations, it is shown also that air PM pollution is higher in Muha than in Mukaza and Ntahangwa municipalities.

\begin{center}
	\begin{figure}[!ht]
		\includegraphics[scale=0.6]{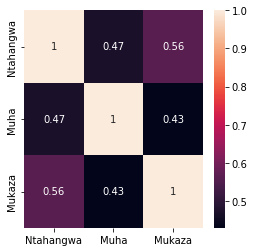}
		\caption{Correlation of PM$_{2.5}$ mass concentrations in Bujumbura municipalities}
	\end{figure}
\end{center}
The higher PM$_{2.5}$ concentrations in Muha compared to other municipalities can be explained by several factors. In Mukaza and part of Ntahangwa, there are public trash cans in some places for better waste management, which is not the case for Muha and that causes waste burning in this locality. Mukaza is largely commercial, part of Ntahangwa is industrial, while Muha is largely residential where wood or charcoal is largely fuel used for cooking. Furthermore, road dust re-suspension can be very important in Muha compared to the other sites given the lower quality of roads networks in this locality. In addition, the Muha commune is bordered with the crop fields of non-urban communes such as Kanyosha and Kabezi, which contributes to air pollution with field waste fires. The time series of the three sites were also compared between them using Pearson correlation as shown in \textbf{Figure 6.} Overall, weak to moderate correlation was observed suggesting that those sites are affected by different sources. Mukaza was the least correlated with other sites indicating different more localized emission sources in that municipality.

\subsection*{3.5. Forecasting PM$_{2.5}$ concentrations using Long Short-Term Memory Model}
Forecasting air pollution in Bujumbura with Long Short-Term Memory Model has been carried out and results are given in \textbf{Figure 7}. PM$_{2.5}$ hourly mean concentrations are used to train the model. Model evaluation is given at each site by  root mean square error and absolute mean error.  The RMSE is in range of 8.2 $\mu$g/m$^3$ to 12.8 $\mu$g/m$^3$ and the MAE in range of 5.3 $\mu$g/m$^3$ to 6.8 $\mu$g/m$^3$. This shows that the recurrent neural network method is a potential forecasting model in Bujumbura for PM$_{2.5}$ time series data. A good correlation is obtained with $r^2$, in the range of 0.7 to 0.8.
\begin{figure}[htp]
\centering 

\subfloat[Mukaza PM$_{2.5}$  forecasting]{%
  \includegraphics[clip,width=0.85\columnwidth]{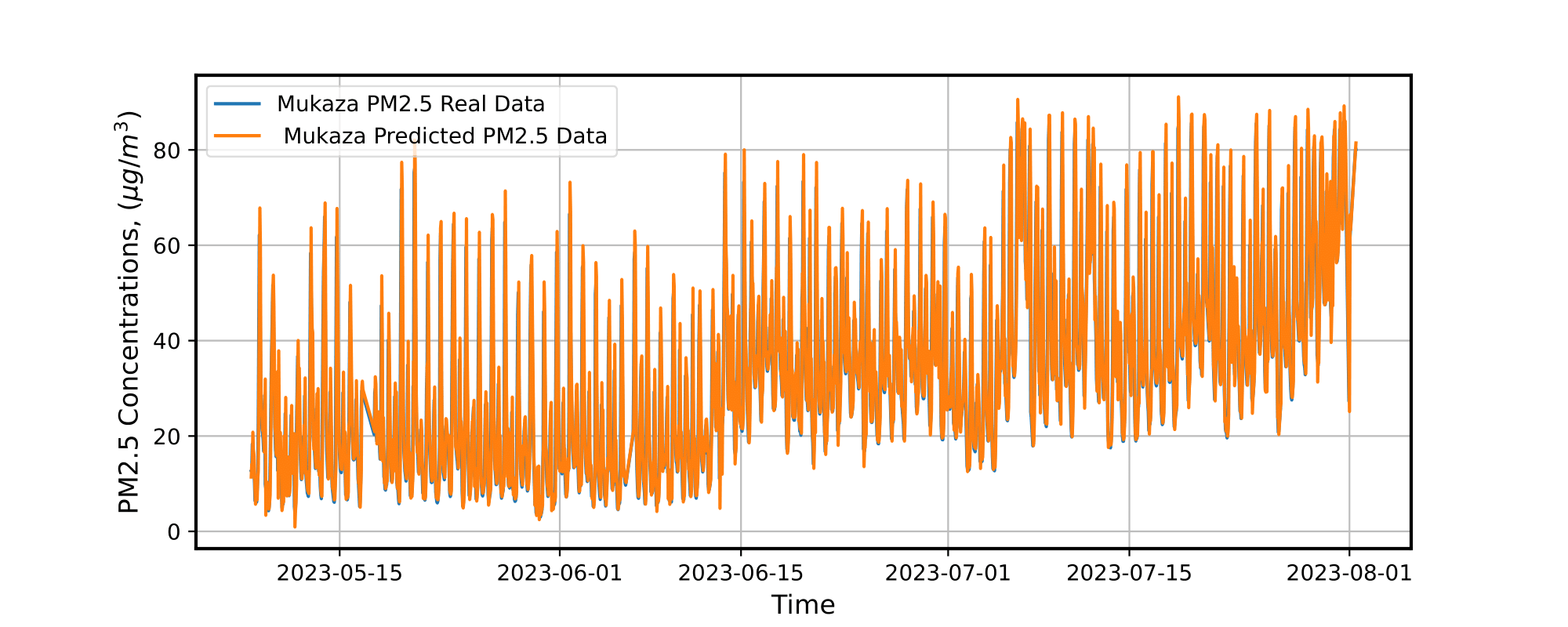}%
}

\subfloat[Muha PM$_{2.5}$ forecasting]{%
  \includegraphics[clip,width=0.85\columnwidth]{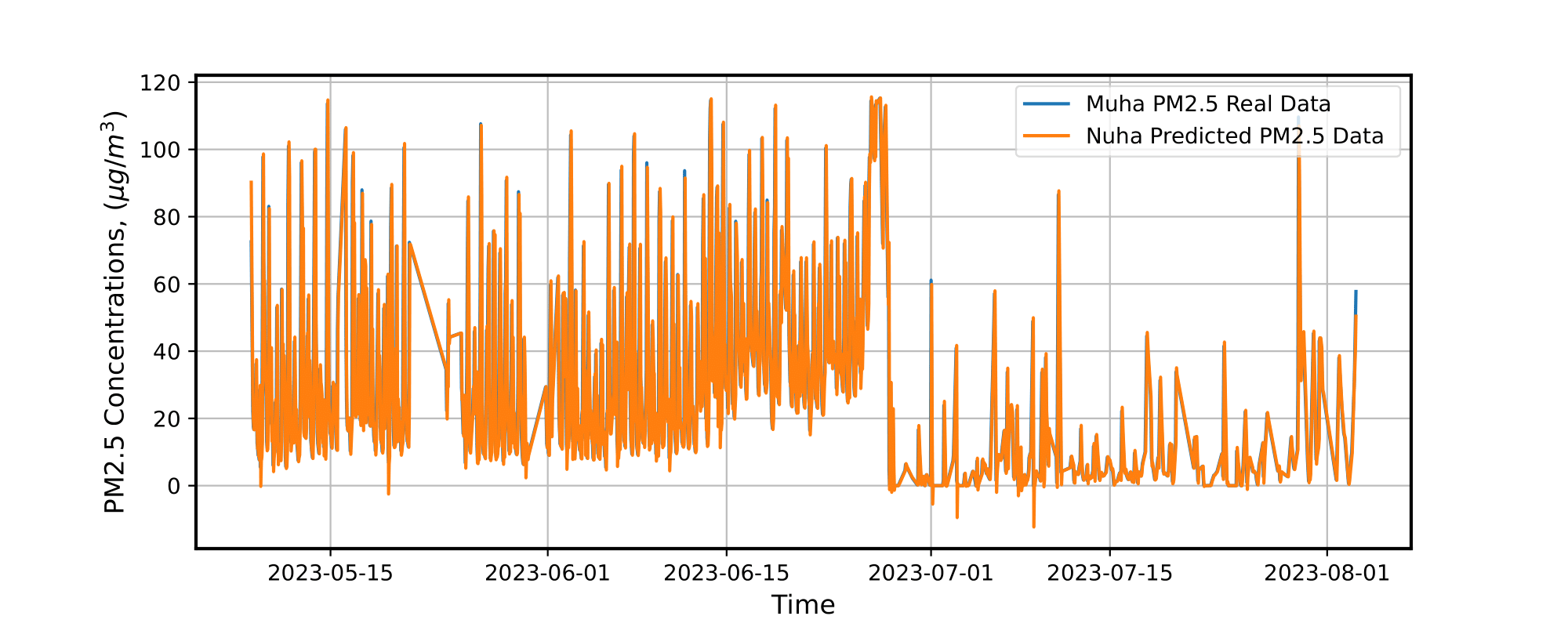}%
}

\subfloat[Ntahangwa PM$_{2.5}$  forecasting]{%
  \includegraphics[clip,width=0.85\columnwidth]{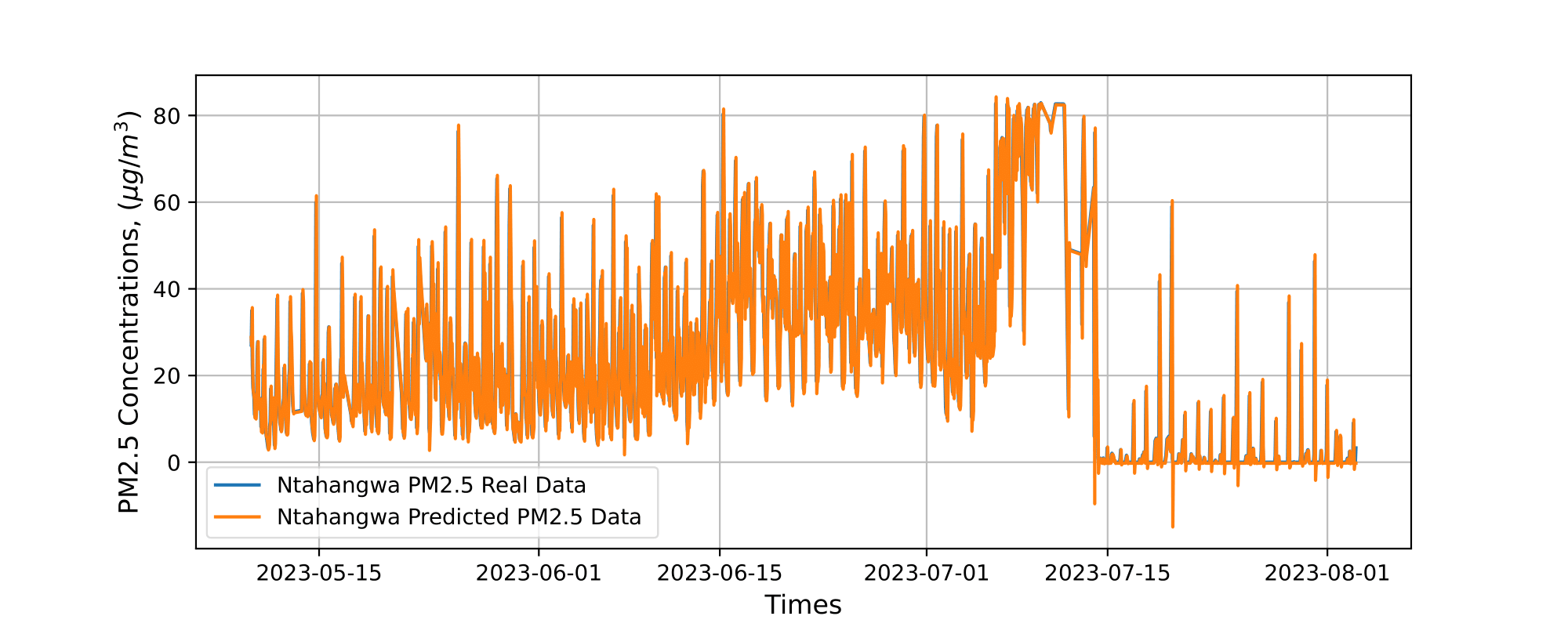}%
}

\caption{Bujumbura PM$_{2.5}$ forecasting }

\end{figure}

\section*{4. Conclusion}
The very first characterization of  spatio-temporal variability of  PM$_{2.5}$ in Bujumbura has been explored. Hourly, daily and seasonal analysis were carried out on the three communes of Bujumbura. The analysis has showed that the PM$_{2.5}$ concentration  is location dependent in the study area. An investigation of Recurrent Neural Networks-Long Short Term Memory has given good performance as a potential forecasting model in Bujumbura.  

\section*{Acknowledgment}
The author, Egide NDAMUZI, would like to thank  the African Institute for Mathematical Sciences, \url{www.nexteinstein.org}, with financial support from the Government of Canada, provided through Global Affairs Canada, \url{www.international.gc.ca}, and the International Development Research Centre, \url{www.idrc.ca} who  funded this work.

\bibliographystyle{ieeetr}
\bibliography{refs}

\end{document}